\documentclass[sigconf]{acmart}

\usepackage{amsmath,amssymb,amsfonts}
\usepackage[ruled,vlined,linesnumbered]{algorithm2e} 
\usepackage[percent]{overpic}
\usepackage{graphicx}
\usepackage{threeparttable} 
\usepackage{textcomp}
\usepackage{xcolor}
\usepackage{booktabs}
\usepackage{stfloats}
\usepackage{placeins}
\usepackage{multirow}
\usepackage{tabularx}
\usepackage{microtype}
\usepackage{caption}
\usepackage{subcaption}
\AtBeginDocument{%
  }

\setcopyright{acmlicensed}
\copyrightyear{2026}
\acmYear{2026}
\setcopyright{cc}
\setcctype{by}
\acmConference[KDD '26]{Proceedings of the 32nd ACM SIGKDD Conference on Knowledge Discovery and Data Mining V.2}{August 09--13, 2026}{Jeju Island, Republic of Korea}
\acmBooktitle{Proceedings of the 32nd ACM SIGKDD Conference on Knowledge Discovery and Data Mining V.2 (KDD '26), August 09--13, 2026, Jeju Island, Republic of Korea}
\acmDOI{10.1145/3770855.3817946}
\acmISBN{979-8-4007-2259-2/2026/08}

\begin{document}

\title{CMF-ELN: A Cross-Modal-Fused End-to-end Learning Network for Cold-Start Drug-Drug Interaction Prediction}

\author{Di Wu}
\affiliation{%
  \institution{College of Computer and Information Science, \\ Southwest University}
  \city{Chongqing}
  \country{China}
}
\email{wudi1986@swu.edu.cn}

\author{Hongyi Sun}
\affiliation{%
  \institution{College of Computer and Information Science, \\ Southwest University}
  \city{Chongqing}
  \country{China}
}
\email{swulhx001@email.swu.edu.cn}

\author{Haichao Xu}
\authornote{Corresponding author.}
\affiliation{%
  \institution{College of Computer and Information Science, \\ Southwest University}
  \city{Chongqing}
  \country{China}
}
\email{popins@email.swu.edu.cn}

\author{Jia Chen}
\affiliation{%
  \institution{Beihang University}
  \city{Beijing}
  \country{China}
}
\email{chenjia@buaa.edu.cn}

\author{Zhong Chen}
\affiliation{%
  \institution{School of Computing, \\ Southern Illinois University Carbondale}
  \city{Carbondale}
  \state{IL}
  \country{United States}
}
\email{zhong.chen@siu.edu}

\author{Jie Yang}
\affiliation{%
  \institution{School of Physics and Electronic Science, \\ Zunyi Normal University}
  \city{Zunyi}
  \state{Guizhou}
  \country{China}
}
\email{2012009@zync.edu.com}


\renewcommand{\shortauthors}{Di Wu et al.}

\begin{abstract}
    Cold-start drug–drug interactions (DDIs) prediction of new drugs is critical for minimizing unexpected adverse drug reactions. The crux of cold-start DDI prediction is to capture the \textit{Similarity} between new and known drugs. However, such similarity is closely associated with complex relationships and mechanisms among drugs, enzymes, transporters, molecular structures, etc. Existing methods have three limitations in capturing such similarity: (1) only partial relationships and mechanisms are considered, which overlooks cross-modal information and yields incomplete or biased similarity modeling; (2) similarity computation between new and known drugs is conducted separately across modalities and performed offline for cold-start DDI prediction, leading to an increasing misalignment between the similarity computation and the DDI prediction stages; (3) existing interpretability analyses are typically single-modality and focus primarily on the key determinants of the perpetrator drug, while the underlying causes of susceptibility for the victim drug are seldom investigated. To address these issues, this paper proposes a novel Cross-Modal-Fused End-to-end Learning Network (CMF-ELN) with three components. First, diverse Multi-modal information is leveraged to construct four types of drug-centered knowledge graphs, enabling comprehensive similarity modeling between new and known drugs under reconstruction based supervision. Second, a four-channel graph autoencoder is designed to fuse cross-modal similarity within an end-to-end learning framework. Finally, a two-stage interpretability scheme is devised to precisely localize the key factors for both the perpetrator and the victim drugs. Extensive experiments on two real datasets demonstrate that CMF-ELN achieves significantly higher prediction accuracy and more comprehensive interpretability of mechanisms than its peers. The source code and datasets are publicly available in our GitHub repository: \url{https://github.com/lhx-cmd/CMF-ELN}.
\end{abstract}




\maketitle

\section{Introduction}
In drug development, predicting DDIs is critical for ensuring therapeutic safety, as co-administration can alter pharmacological properties and lead to severe adverse reactions \cite{DeepDDI, wienkers2005predicting, percha2013informatics}. Since traditional experimental assays are often cost-prohibitive and time-consuming to keep pace with new drug discovery, computational approaches have emerged as efficient alternatives \cite{DDIMDL, vilar2014similarity}. Notably, recent deep learning models demonstrate significant scalability by integrating chemical and biological information to automatically uncover latent drug relationships \cite{DPDDI2020, huang2020deeppurpose, chen2020transformercpi,MKG-FENN,Zhao2025RegulationDrugRepositioning,Qiao2025TCMHypergraph,Wu2025MiRNADrug}.

Among various DDI prediction tasks, cold-start DDI prediction is particularly challenging \cite{huang2020caster}. This task aims to predict potential interactions between newly developed drugs and known drugs, where the new drugs usually lack any known DDI records and often exhibit modality missing or sparsity (e.g., incomplete
enzymes, transporters, or structural annotations). Therefore, the model must infer potential interactions based on structural and biological \textit{Similarities} between drugs while adapting to incomplete multimodal evidence. This requirement is closely related to representation learning under sparse, incomplete, and high-dimensional observations, where latent factor, autoencoding, and tensor-learning models have been widely explored \cite{Wu2021RobustLFA,Wu2024MMLF,Wu2025OutlierAutoencoder,Wu2026MMAutoencoder,Wu2026NGHFL}. Achieving this goal requires leveraging heterogeneous and multimodal information sources which jointly form a complex multimodal knowledge network encoding multiple aspects of drug-related mechanisms \cite{yu2021sumgnn, ma2018drug, Zhang2022,Wu2026FLFT,Yu2026FLF,Tang2026MPSANT}. Prior multimodal DDI frameworks highlight the benefit of integrating complementary signals \cite{DDIMDL}. Substructure-level interactions further enrich the structural perspective \cite{SSI-DDI}. Comprehensive drug knowledge bases also provide essential annotations and links across modalities \cite{DrugBank2018}.

Despite recent progress, existing methods still face three critical challenges in capturing similarity between new and old drugs for cold-start DDI prediction.
First, many approaches only consider partial relationships or mechanisms (e.g., a single molecular \cite{DeepDDI, SSI-DDI, GCNMK2022, huang10molecular, zitnik2018modeling} or biomedical view \cite{KGNN, DDKG, xiong2019pushing}), overlooking cross-modal correlations that are essential for faithful similarity modeling. 
Second, similarity between new and old drugs is frequently computed separately across modalities and offline \cite{GCNMK2022, Zhang2022, Tatonetti2012, gottlieb2012indi, rohani2019drug, amiri2022novel}, and then treated as fixed inputs to a downstream predictor. 
Third, interpretability analyses for DDI prediction are typically single-modality and focus primarily on determinants of the perpetrator drug, while the mechanistic causes underlying the victim drug's susceptibility are seldom examined \cite{MeTDDI2024,tang2023dsil,wang2024zeroddi}. 

To address these challenges, we propose the Cross-Modal-Fused End-to-end Learning Network (CMF-ELN) for cold-start DDI prediction. CMF-ELN operates on a unified principle: maximizing the utilization of cross-modal biomedical evidence to enable robust similarity learning and mechanistic transparency. The specific contributions are organized as follows: First, to construct comprehensive feature representations, the framework systematically extracts information from proteins, Morgan fingerprints, MACCS keys, and structural motifs to build four distinct knowledge graph channels. Second, these channels are integrated into a graph autoencoder that unifies representation learning and DDI prediction within a single end-to-end pipeline, ensuring seamless fusion of cross-modal data. Finally, CMF-ELN transcends the limitations of traditional single-sided analysis by incorporating a bidirectional interpretability mechanism. By precisely localizing key risk factors for both the perpetrator and the victim drugs, our model provides a comprehensive explanatory logic chain and actionable insights to assist medical personnel in proactive risk evaluation.

We summarize our main contributions as follows:
\begin{itemize}
    \item \textbf{CMF-ELN Model.} We propose CMF-ELN to achieve
    high-accuracy cold-start DDI prediction and provide comprehensive DDI mechanism explanations.

    \item \textbf{Multi-Modal Drug Knowledge Graph Construction.} We devise a multi-modal drug knowledge construction strategy that builds four distinct drug-centered knowledge graphs, leveraging diverse biomedical data to enable comprehensive similarity modeling for cold-start drugs.

    \item \textbf{End-to-End Integrative Joint Optimization.} We propose a four-channel graph autoencoder that jointly optimizes similarity learning and DDI prediction. By replacing offline computations with an end-to-end approach, the framework aligns multi-scale embeddings and enhances robustness against noisy or missing data in cold-start scenarios.

    \item \textbf{Dual-End Interpretability.} We introduce a two-stage scheme that identifies dominant biological modalities and attributes key features. This uniquely reveals the "perpetrator" aggressive factors and "victim" vulnerabilities, providing transparent mechanistic insights for clinical safety and pharmacological research.
\end{itemize}

\section{Related Work}

\subsection{Molecular Structure-Based Methods}
Motifs, defined as significant subgraph patterns frequently occurring within molecular structures \cite{Milo2002}, have emerged as a powerful paradigm for enhancing molecular representation learning. Recent studies have leveraged motif-based approaches to capture intricate structural and functional relationships in molecular graphs. For instance, Zhang et al. \cite{Zhang2021} propose a motif-based pre-training framework (MGSSL) to extract multiscale information from molecular graphs, improving the robustness of molecular representations. Similarly, SSI-DDI \cite{SSI-DDI} decomposes the DDI prediction task into identifying pairwise interactions between substructures (motifs), enabling a more granular analysis of molecular interactions. Further advancing this paradigm, Yu and Gao \cite{Yu2022} construct a heterogeneous graph incorporating motif and molecular nodes to model motif-level relationships, enhancing the understanding of complex molecular interactions via their HM-GNN model. Additionally, Ye et al. \cite{Ye2024} introduce a hierarchical cross-level graph contrastive learning framework that extracts semantic motifs and captures intricate connections within drug-motif interaction graphs.

Despite these advancements, relying on a single structural modality often results in limited generalizability for cold-start DDI prediction. These models struggle to capture the latent pharmacological properties of new drugs that are not fully captured by visible subgraph patterns, leading to a lack of robust discriminative signals when encountering novel chemical entities.

\subsection{Knowledge Graph-Based Methods}
Knowledge Graphs (KGs) facilitate a comprehensive approach to DDI modeling by integrating diverse biological entities and their relationships, such as drugs, targets, enzymes, and transporters. Several methodologies leverage KGs to enhance DDI prediction through advanced graph-based neural architectures. For instance, KGNN \cite{KGNN} employs Graph Convolutional Networks (GCNs) combined with neighborhood sampling to effectively capture valuable relational information within the graph structure. Similarly, La-GAT \cite{LaGAT} and HAN-DDI \cite{tanvir2023predicting} utilize link-aware and heterogeneous graph attention mechanisms respectively, generating multiple attention pathways for drug entities based on the heterogeneous connections between drug pairs. Furthermore, KGE-NFM \cite{ye2021unified} introduces a unified framework combining knowledge graph embeddings with neural factorization machines to model explicit and implicit feature interactions among drugs and their biological neighbors. DDKG \cite{DDKG} advances these efforts by learning drug embeddings from their inherent attributes within KGs, simultaneously incorporating neighboring node embeddings and relational triples to enhance representation quality. Emer-GNN \cite{EmerGNN} focuses on predicting interactions for emerging drugs by exploiting the rich contextual information embedded in biomedical networks. Additionally, KnowDDI \cite{KnowDDI} refines drug representations by adaptively leveraging extensive neighborhood information from large-scale biomedical KGs. 

However, the lack of cross-modal integration between molecular structures and KG relations often results in suboptimal performance for cold-start DDI prediction, as these independent information streams fail to provide a holistic biological representation.

\subsection{Interpretation for DDI Mechanisms}
Enhancing the interpretability of DDI prediction is pivotal for elucidating the pharmacological mechanisms underlying interactions. A prominent line of research has focused on refining structural attention to localize risk factors. For instance, MetDDI \cite{MeTDDI2024} utilized local-global self-attention to capture motif-level interactions. Most recently, Mrhgnn \cite{chen2025towards} advanced this direction by establishing a hierarchical framework that provides explanations at both molecular and network levels. Parallel to structural analysis, addressing interpretability in out-of-distribution or cold-start scenarios has gained significant traction. DSIL-DDI \cite{tang2023dsil} proposed domain-invariant substructure learning to extract causal, generalizable interaction patterns across different distributions. Building on the need for robust generalization, ZeroDDI \cite{wang2024zeroddi} further integrated semantic descriptions with molecular graphs, enabling logic-based zero-shot reasoning for unseen drugs. 

Despite providing local insights, these methods lack a multi-modal explanatory perspective, failing to identify the contributions of diverse biological views or characterize the interaction mechanisms from the dual perspectives of perpetrator and victim drugs.

\subsection{Representation Learning for Incomplete and Heterogeneous Data}
The cold-start setting in DDI prediction shares core challenges with broader studies on sparse, incomplete, and heterogeneous data representation. A series of latent factor and feature-selection models by Wu, Luo, and collaborators investigate robust representations for high-dimensional sparse matrices, online streams, and web-service quality prediction \cite{Wu2018SelfTrainingDP,Wu2018SelfLabeled,Wu2019DataAwareLFM,Wu2019CapriciousFS,Wu2020PMLF,Wu2021DeepLFM,Wu2021RobustLFA,Wu2022DataCharacteristicLFM,Wu2022PosteriorLFM,Wu2022L1L2LFM,Wu2022OnlineSparseFS,Wu2023DSDNLF,Wu2023GraphLFA,Wu2024MMLF,Wu2024PSMLF,Wu2025RobustLowRankSignal,Wu2025OutlierAutoencoder,Wu2026MMAutoencoder,Wu2026NGHFL,Xu2025RFFeatureSelection,Xu2025ThreeWayFeatureSelection,Luo2019UnconstrainedNLFA,Yuan2020FastNLF}.
Recent tensor, federated, and autoencoding extensions further emphasize learning reliable latent structure from dynamic, privacy-sensitive, and partially observed data \cite{Wu2026FLFT,Yu2026FLF,Yuan2026TemporalQoS,Tang2026MPSANT,Hou2026MASANT,Li2026AdaptivePIDNFA,He2026TensorCompressionSurvey,He2026TensorLowRankCNN,Ma2025PowerLoadSurvey,Yu2025MultiIndicatorTensor,Gao2025FederatedDeepLFM,Wu2025ModeAwareTucker,Wang2025ConvBiasTensor,Liao2025ProxADMMTensor,Liao2025TensorCausalConv,Tang2025AutoEncodingNTF,Lin2025NNLatentFactor,Xu2025AttentionNTF,Tang2024TemporalQoS,Yuan2024FuzzyPIDSGD,Chen2024StateMigrationPSO,Zhong2024ADMMNLFA,Wang2024DistributedLFA,Chen2024GNAGNLFT,Lyu2025GeneticLFA}.
In parallel, graph neural representation learning has expanded from efficient neighborhood aggregation to high-order, dual-channel, dynamic, and message-passing formulations \cite{He2026ModularGCN,Wang2026HighOrderGCN,Lin2026DualChannelGCN,Chen2025KHopGCN,Gou2025MCDGNN,Lin2025NCSAC,Han2025SGDDYG,Chen2025NeighborhoodGCN,Bi2024GraphLinearPooling,Yuan2024NodeCollaborationGCN,Chen2024SDGNN,He2024PolarizedMPNN,Yang2024CognitiveGraph,Zhu2024SequenceDenoising,He2024StructureSelfAttention,Wu2026SchemaRAG,Yang2025GraphClustering}. These directions also connect to biomedical and healthcare network learning, including drug repositioning, miRNA--drug association prediction, lncRNA--miRNA interaction modeling, protein-complex identification, and pathological grading \cite{Zhao2025RegulationDrugRepositioning,Qiao2025TCMHypergraph,Wu2025MiRNADrug,Zhao2024LncRNAMiRNA,Yang2025FMvPCI,Li2025KnowledgeDrivenMIL}. These studies motivate CMF-ELN's focus on fusing complementary modalities while explicitly handling sparsity and missingness in cold-start DDI prediction.


\section{Preliminaries}

\textbf{Drug Knowledge Graph.}
The drug knowledge graph, denoted as $G = (D, T, R)$, is a specialized knowledge graph constructed for the purpose of predicting DDI. It comprises three primary components: $D$, representing a subset of drug entities; $T$, representing a subset of tail entities associated with drugs (e.g., drug-metabolizing enzymes); and $R$, denoting the set of relations between drugs and tail entities. Formally, the drug knowledge graph is defined as a collection of triples $(d, t, r_{dt})$, where each triple describes a connection between a drug entity $d \in D$, a tail entity $t \in T$, and a relation $r_{dt} \in R$. Systematic analysis of $G$ yields valuable insights into the relationships between drugs and their associated tail entities, thereby facilitating more accurate prediction of DDI.

\textbf{Cold-start DDI Prediction.}
Our primary objective is to predict specific DDI between a given pair of drugs $d_i$ and $d_j$ by leveraging both the DDI matrix $\mathbf{Y}$ and the drug knowledge graph $G$. To this end, we define the prediction function $\hat{y}_{ij} = \Gamma\!\left( d_i, d_j \,\middle|\, \Theta, \mathbf{Y}, G \right)$, where $\Theta$ denotes the set of model parameters and $\Gamma(\cdot)$ integrates information from $\mathbf{Y}$ and $G$ to produce an estimate $\hat{y}_{ij}$ representing the likelihood or category of the interaction between $d_i$ and $d_j$. By jointly considering heterogeneous sources of information and model parameters, this formulation aims to improve both the accuracy and robustness of DDI prediction.

\section{Proposed Method}
\begin{figure*}[t] 
    \centering 
    \includegraphics[width=0.95\textwidth]{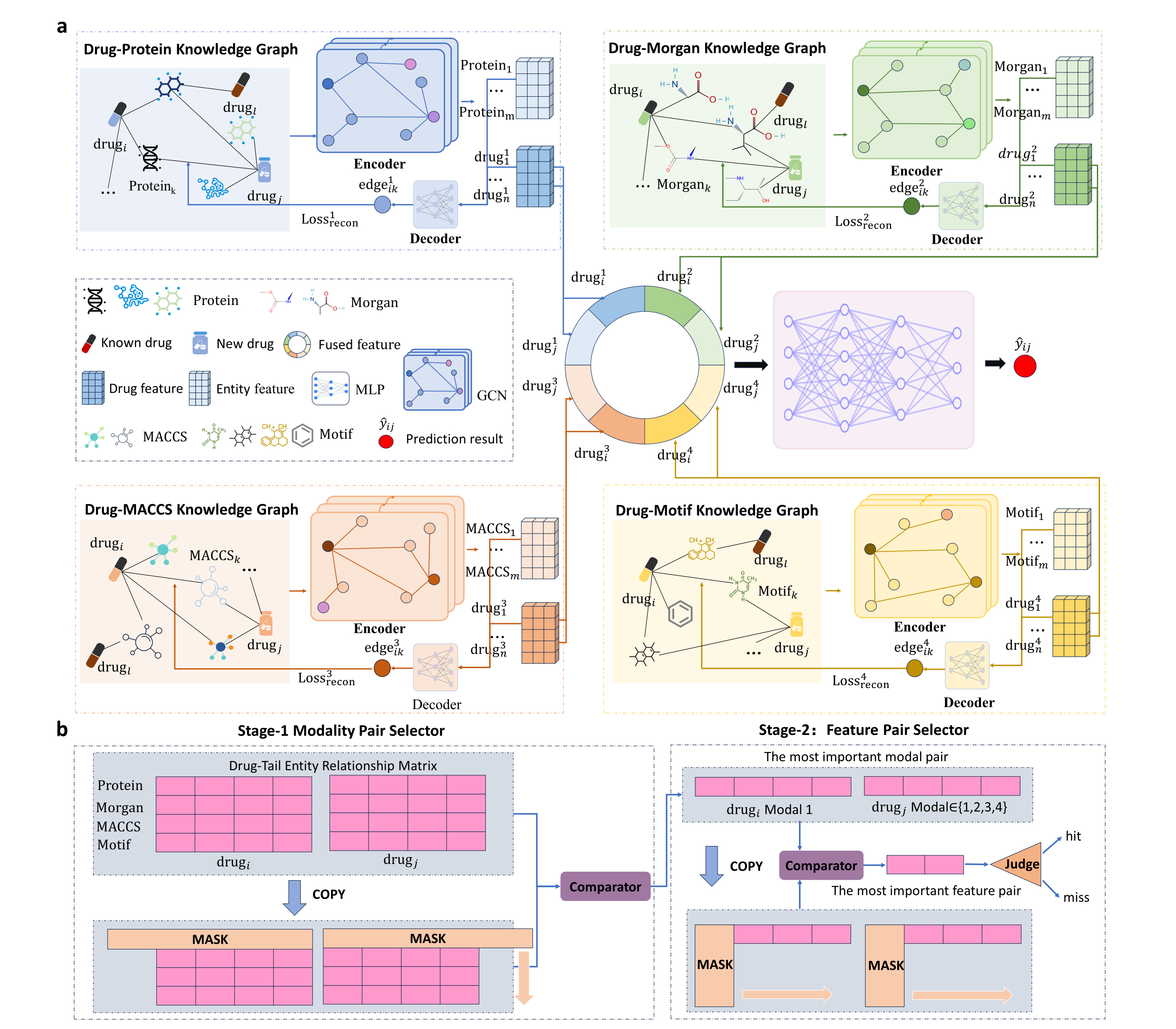} 
    \caption{(a) The overall structure of the proposed CMF-ELN model. (b) Two-stage interpretability algorithm schematic for localizing key feature pairs.
} 
    \label{fig:over_structure} 
\end{figure*}

\textbf{Overview.} 
The overall framework of the proposed CMF-ELN is illustrated in Fig.~\ref{fig:over_structure}. Specifically, Fig.~\ref{fig:over_structure}(a) depicts the end-to-end learning pipeline for cross-modal fusion and DDI prediction, while Fig.~\ref{fig:over_structure}(b) presents the bidirectional interpretability mechanism designed to localize key risk factors. The following subsections describe these components in detail, outlining the specific methodologies employed in each module.

\subsection{Multi-Modal Drug Knowledge Graph Construction}

To comprehensively capture the multifaceted characteristics of drugs, we construct four distinct modality-specific knowledge graphs, each encoding a specific type of biological or chemical information. Each graph comprises triples formatted as $\langle \text{drug}, \text{entity}, \text{relation} \rangle$, wherein \text{entity} represents a distinct biological component or structural attribute, and \text{relation} signifies the inclusion or interaction. The construction of each component graph is detailed below:

\textbf{Drug-Protein.} We utilize DrugBank\cite{DrugBank2018} to extract proteins associated with DDIs. Specifically, we focus on two critical categories: drug-metabolizing enzymes and drug transporters. Enzymes catalyze biotransformation, whereas transporters regulate drug uptake and efflux. Accordingly, we construct triples of the form $\langle \text{drug}, \allowbreak \text{protein}, \allowbreak \text{relation} \rangle$, indicating the specific interaction type (e.g., metabolic involvement or transporter-mediated binding).

\textbf{Drug-Morgan.} We generate Morgan fingerprints\cite{rogers2010extended} to capture the local topological structures of drugs. Specifically, for each drug molecule, the algorithm iterates through each atom to encode its immediate neighborhood within a defined radius into a fixed-length binary vector. We then convert the activated fingerprint bits from their binary positions to decimal indices, which are utilized as discrete fingerprint-bit identifiers. Accordingly, we construct triples of the form $\langle \text{drug}, \allowbreak \text{fingerprint-bit index}, \allowbreak \text{include} \rangle$, indicating that the drug contains the corresponding Morgan fingerprint bit.

\textbf{Drug-MACCS.} The MACCS-key protocol follows the standard definition~\cite{durant2002reoptimization}. For each drug, we compute its MACCS fingerprint and convert activated bits to decimal indices as discrete identifiers. In this work, we retain 13 selected MACCS keys as drug-associated entities. Accordingly, we construct triples $\langle \text{drug}, \allowbreak \text{index}, \allowbreak \text{include} \rangle$, indicating the presence of a MACCS key in the drug.

\textbf{Drug-Motif.} Motifs are drug molecular substructures obtained by fragmenting each molecule according to chemistry-informed reaction rules\cite{degen2008art}. Accordingly, we construct triples of the form $\langle \text{drug}, \allowbreak \text{motif}, \allowbreak \text{include} \rangle$, indicating that the motif is a fragmented substructure derived from the drug.

By establishing these four parallel knowledge graphs, the system provides diverse drug-related views to support the subsequent cross-modal learning. This multi-view representation effectively compensates for the sparsity or absence of specific modality data in certain drugs, establishing a robust foundation for the feature extraction and DDI inference within the CMF-ELN framework.

\subsection{End-to-End Integrative Joint Optimization}
\subsubsection{GCN-based Graph Autoencoder}

To effectively capture the structural and semantic information within the cross-modal knowledge graph, we employ a Graph Autoencoder (GAE) architecture based on Graph Convolutional Networks (GCN), following the broader practice of using graph encoders and masked or autoencoding objectives for robust representation learning \cite{Hou2022,He2026ModularGCN,Chen2025KHopGCN,He2024PolarizedMPNN}.
Structurally, this framework is composed of two coupled modules: an Encoder and a Decoder. The Encoder is designed to project heterogeneous node features into a shared low-dimensional latent space via graph convolution operations. Conversely, the Decoder operates on these latent embeddings to reconstruct the topological relations between entities, thereby supervising the model to learn expressive and robust representations. The specific designs of these two components, the Encoder and the Decoder, are detailed as follows.

\noindent\textbf{Encoder for Feature Representation.}
Serving as the encoding component, this module maps the heterogeneous graph data into a shared latent space. Specifically, to effectively capture both the semantic attributes and topological structure within the cross-modal knowledge graph, we employ a GCN-based architecture. This encoder transforms high-dimensional node features into compact low-dimensional embeddings ($\mathbf{Z}$) while preserving local graph neighborhoods. The detailed framework involves feature projectors and stacked graph convolutional layers, as elaborated below:

We differentiate between two node categories: drug nodes and tail-entity nodes. Let $\mathbf{X}_{\mathrm{drug}}$ and $\mathbf{X}_{\mathrm{tail}}$ denote their initial feature representations with dimensions $N_d \times F_d$ and $N_t \times F_t$, respectively. To align these heterogeneous feature spaces, we employ two type-specific feature transformation modules to map the drug and tail features into a shared latent space with dimension $F$. Both modules share an identical architecture, parameterized as a two-layer MLP equipped with normalization, ReLU activation, and dropout. The transformed features are subsequently concatenated to form a unified node feature matrix $\mathbf{X}\in\mathbb{R}^{(N_d+N_t)\times F}$ for the subsequent graph convolution layers. Formally, the layer-wise propagation within these modules is defined as:
\begin{equation}
\mathbf{H}^{(l+1)}=\sigma\!\left(\mathrm{Norm}\!\left(\mathbf{H}^{(l)}\mathbf{W}^{(l)}+\mathbf{b}^{(l)}\right)\right),
\end{equation}
where $\mathbf{H}^{(0)}$ represents the input features (either $\mathbf{X}_{\mathrm{drug}}$ or $\mathbf{X}_{\mathrm{tail}}$), and $\mathbf{W}^{(l)}$ and $\mathbf{b}^{(l)}$ are learnable parameters. The normalization strategy is layer-dependent: LayerNorm is applied after the first linear transformation, while BatchNorm is applied after the second, and $\sigma(\cdot)$ denotes the ReLU activation function.

Then, the $l$-th GCN layer updates node representations via neighborhood aggregation:
\begin{equation}
\mathbf{H}^{(l+1)}=\sigma\!\left(\tilde{\mathbf{D}}^{-\frac{1}{2}}\tilde{\mathbf{A}}\tilde{\mathbf{D}}^{-\frac{1}{2}}\mathbf{H}^{(l)}\mathbf{W}^{(l)}+\mathbf{b}^{(l)}\right).
\end{equation}
Here, $\tilde{\mathbf{A}}=\mathbf{A}+\mathbf{I}$ denotes the adjacency matrix with added self-connections, where $\mathbf{A}$ is the original adjacency matrix and $\mathbf{I}$ is the identity matrix. $\tilde{\mathbf{D}}$ represents the diagonal degree matrix with entries $\tilde{\mathbf{D}}_{ii}=\sum_j \tilde{\mathbf{A}}_{ij}$.
$l\in\{0,1\}$ denotes the layer index in our two-layer GCN, and $\mathbf{W}^{(l)}, \mathbf{b}^{(l)}$ are learnable parameters.
In our implementation, the first GCN layer maps $F$-dimensional inputs to $2D$ hidden dimensions, and the second GCN layer maps $2D$ to $2D$:
$\mathbf{W}^{(0)}\in\mathbb{R}^{F\times 2D}$ and $\mathbf{W}^{(1)}\in\mathbb{R}^{2D\times 2D}$.

Finally, a linear projection produces the $D$-dimensional node embeddings:
\begin{equation}
\mathbf{Z}=\mathbf{H}^{(2)}\mathbf{W}_{\mathrm{out}}+\mathbf{b}_{\mathrm{out}},
\end{equation}
where $\mathbf{W}_{\mathrm{out}}\in\mathbb{R}^{2D\times D}$ and $\mathbf{Z}\in\mathbb{R}^{N\times D}$.

\noindent\textbf{Decoder for Relation Reconstruction.}
In the decoding phase, we seek to recover the original graph structure from the low-dimensional latent space. To achieve this, we employ a decoder module designed to reconstruct the semantic relations between nodes using the learned embeddings $\mathbf{Z}$.
Specifically, for edge-level tasks, we first construct the edge feature for each directed edge $(i,j)$ by concatenating the embeddings of its incident nodes. Formally,
\begin{equation}
\mathbf{e}_{ij} = \big[\, \mathbf{z}_i \mathbin{\|} \mathbf{z}_j \,\big],
\end{equation}
where $\mathbf{z}_i$ and $\mathbf{z}_j$ denote the source and destination node embeddings extracted by the encoder, and $\mathbin{\|}$ denotes vector concatenation. This edge feature is fed into a linear edge classifier, serving as the decoding function, to produce relation logits:
\begin{equation}
\mathbf{u}_{ij}=\mathbf{W}_{\text{edge}}\mathbf{e}_{ij}+\mathbf{b}_{\text{edge}}.
\end{equation}
Here, $\mathbf{W}_{\text{edge}}$ and $\mathbf{b}_{\text{edge}}$ are learnable parameters, and $\mathbf{u}_{ij}$ denotes the predicted relation scores for edge $(i,j)$, effectively mapping the latent representations back to the relation space.

To explicitly guide the learning of relation semantics, we employ a reconstruction-based objective. Specifically, for the $k$-th modality, the edge-level loss $\mathcal{L}_{\text{edge}_k}$ supervises the prediction of the ground-truth relation label between the drug and the tail entity. It is defined as the multi-class cross-entropy loss over the calculated logits:

\begin{equation}
\mathcal{L}_{\text{edge}_k}
= \frac{1}{N_{\text{edge}}^{(k)}}\sum_{m=1}^{N_{\text{edge}}^{(k)}}
\left(
-\log \frac{\exp\!\left(u^{(k)}_{m,r^{(k)}_m}\right)}{\sum_{c=1}^{C_k}\exp\!\left(u^{(k)}_{m,c}\right)}
\right),
\end{equation}
where $N_{\text{edge}}^{(k)}$ denotes the number of supervised drug–tail edges in the $k$-th modality and $C_k$ is the number of relation types. The term $\mathbf{u}^{(k)}=[u^{(k)}_{m,c}]\in\mathbb{R}^{N_{\text{edge}}^{(k)}\times C_k}$ represents the edge logits from Eq. (5) for the $m$-th edge in this modality, while $\mathbf{r}^{(k)}=\{r^{(k)}_m\}$ denotes the ground-truth relation labels.

In summary, the Encoder effectively integrates node feature information and graph topological structure through GCN operations, producing informative node embeddings and edge features optimized via the aforementioned edge-level supervision.

\subsubsection{Multilayer Perceptron Inference Network}

To integrate the cross-modal embeddings generated by the graph autoencoder modules and perform DDI prediction, we design a multilayer perceptron (MLP) inference network. This network fuses drug embeddings from multiple modalities and outputs the DDI prediction results. The architecture and inference procedure are described as follows:

The MLP inference network comprises stacked fully connected layers with non-linear activation, batch normalization, and dropout regularization. Its primary function is to progressively reduce the dimensionality of fused features while preserving discriminative information for DDI prediction.

The input to the MLP network is a fused feature vector constructed from drug embeddings derived from multiple graph autoencoders. For a drug pair $(i, j)$, the fusion process involves two stages: intra-drug fusion (combining embeddings of a single drug from multiple sources) and drug-pair fusion (combining the fused representations of the two drugs).

\begin{enumerate}
    \item \textbf{Embedding Extraction}: For the drug pair $(i, j)$, embeddings are extracted from four graph autoencoding modules (GAE$_1$ to GAE$_4$). Let $\mathbf{z}_i^{(m)} \in \mathbb{R}^C$ and $\mathbf{z}_j^{(m)} \in \mathbb{R}^C$ denote the embeddings of drug $i$ and drug $j$ from the $m$-th GAE, where $C$ is the embedding dimension and $m\in\{1,2,3,4\}$.

    \item \textbf{Intra-Drug Fusion}: For each drug, embeddings from the four GAE modules are concatenated to form a unified node representation. For drug $i$, this is defined as:
    \begin{equation}
    \mathbf{h}_{i}
    = \big[\, \mathbf{z}_i^{(1)} \mathbin{\|} \mathbf{z}_i^{(2)} \mathbin{\|} \mathbf{z}_i^{(3)} \mathbin{\|} \mathbf{z}_i^{(4)} \,\big],
    \end{equation}
    where $\mathbf{h}_{i} \in \mathbb{R}^{4C}$. Similarly, the fused representation of drug $j$, denoted as $\mathbf{h}_{j} \in \mathbb{R}^{4C}$, is obtained, ensuring symmetric feature structures.
    
    \item \textbf{Drug-Pair Fusion}: The fused representations of the two drugs are concatenated to form the final feature vector for the drug pair, i.e., $\mathbf{v}_{ij} = \big[\, \mathbf{h}_{i} \mathbin{\|} \mathbf{h}_{j} \,\big]$. Here, $\mathbf{v}_{ij}\in\mathbb{R}^{8C}$ integrates the cross-modal information from both drugs.

    \item \textbf{Inference via MLP}: The fused vector $\mathbf{v}_{ij}$ is processed through three stacked linear layers, each applying a linear transformation with bias, followed by ReLU activation and batch normalization. The operation of the $(l{+}1)$-th layer is defined as:
    \begin{equation}
    \mathbf{H}^{(l+1)}=\sigma\!\left(\mathbf{H}^{(l)}\mathbf{W}^{(l)}+\mathbf{b}^{(l)}
    \right),
    \end{equation}
    where $l\in\{0,1,2\}$, and the input is initialized as $\mathbf{H}^{(0)}=\mathbf{v}_{ij}$. The final layer outputs the prediction logits $\hat{\mathbf{y}}_{ij}$, where the output dimension corresponds to the total number of DDI types, denoted as $K$.
    
\end{enumerate}

To optimize the model for accurate DDI event classification, we employ a supervised learning objective based on the model outputs. Specifically, the label-level loss, $\mathcal{L}_{\text{label}}$, is computed using the standard multi-class cross-entropy function. Considering a mini-batch of $N$ samples, let $\mathbf{y}_n \in \{0, 1\}^{D}$ denote the one-hot encoded ground-truth vector for the $n$-th drug pair, and $\hat{\mathbf{y}}_n \in \mathbb{R}^{D}$ denote the predicted raw logits. The loss is defined as:

\begin{equation}
\mathcal{L}_{\text{label}} = - \frac{1}{N} \sum_{n=1}^{N} \sum_{d=1}^{D} y_{n,d} \log \left( \frac{\exp(\hat{y}_{n,d})}{\sum_{j=1}^{D} \exp(\hat{y}_{n,j})} \right),
\end{equation}
where $D$ represents the total number of DDI types, $y_{n,d}$ is the binary indicator ($0$ or $1$) confirming if class $d$ is the correct label for sample $n$, and $\hat{y}_{n,d}$ represents the output logit for the corresponding class.

\subsubsection{Joint Supervision with Label Prediction and Edge Reconstruction}

To jointly optimize the GAE-based representation learning and the MLP-based inference, we formulate a total objective function by aggregating the previously defined loss terms. This composite loss balances the primary DDI classification task with the auxiliary edge reconstruction tasks, and is expressed as:
\begin{equation}
\mathcal{L} = \mathcal{L}_{\text{label}} + \sum_{k=1}^{4} \lambda_k \,\mathcal{L}_{\text{edge}_k},
\end{equation}
where $\lambda_k$ are hyperparameters that control the relative contribution of the reconstruction loss $\mathcal{L}_{\text{edge}_k}$ for the $k$-th modality.

By minimizing this unified objective, the model performs end-to-end learning of embeddings that not only reconstruct modality-specific graph structures but also align with the DDI prediction goal. This coupling mechanism induces pair-specific, context-conditioned similarity between drugs, enabling the model to adapt to missing or noisy modalities while strengthening generalization performance.

\subsection{Dual-End Interpretability}

In line with MeT-DDI\cite{MeTDDI2024}, we use the same interpretability evaluation data constructed under its scenario definition. Specifically, MeT-DDI motivates explanations by a mechanistic chain in which $\text{drug}_j$ inhibits the activity of the metabolic enzymes responsible for $\text{drug}_i$, thereby reducing $\text{drug}_i$'s metabolic rate. Following this rationale, and given our multi-modal evidence space, we fix the victim-side evidence of $\text{drug}_i$ to modality~1 (enzyme-related features), while allowing $\text{drug}_j$ to select its most influential modality from $\{1,2,3,4\}$.

To interpret the mechanistic evidence underlying the model's predictions, we employ a hierarchical perturbation-based attribution scheme. By masking specific edges in the input relation matrix and quantifying the resulting shift in the fused representation, we identify critical evidence in a coarse-to-fine manner (Fig.~\ref{fig:over_structure}(b)): Stage~1 determines the dominant modality for $\text{drug}_j$, which restricts the search space for Stage~2 to localize the decisive tail-entity pair.

We quantify perturbation by the squared $\ell_2$ distance. Let $\mathbf{R}_{\text{full}}$ denote the complete relation matrix characterizing the connections between drugs and tail entities, and let $f(\cdot)$ represent the trained fusion model that maps this relation matrix to the final fused representation. Let $m_i$ and $m_j$ denote the selected modalities for $\text{drug}_i$ and $\text{drug}_j$, respectively.

In Stage-1, we fix $m_i=1$ and enumerate $m_j\in\{1,2,3,4\}$. For each $(m_i,m_j)$, we mask the rows in $\mathbf{R}_{\text{full}}$ corresponding to the tail features of $(m_i,\text{drug}_i)$ and $(m_j,\text{drug}_j)$ to form the perturbed matrix $\mathbf{R}^{(m_i,m_j)}_{\text{mask}}$, and define the modality-level perturbation as
\begin{equation}
d(m_i,m_j)=\Big\|f(\mathbf{R}_{\text{full}})-f\!\left(\mathbf{R}^{(m_i,m_j)}_{\text{mask}}\right)\Big\|^{2}.
\label{eq:cs_stage1_dmamb}
\end{equation}
Since larger masking scope tends to induce larger perturbations, we calibrate by the masking size, defined as
\begin{equation}
n_t = \left|\mathrm{tails}^{(1)}_{i}\right| + \left|\mathrm{tails}^{(m)}_{j}\right|.
\label{eq:cs_nt}
\end{equation}
Here, $\left|\mathrm{tails}^{(1)}_{i}\right|$ and $\left|\mathrm{tails}^{(m)}_{j}\right|$ denote the numbers of tail features (i.e., tail entities) associated with $\text{drug}_i$ in modality 1 and $\text{drug}_j$ in modality $m$, respectively; thus $n_t$ measures the total number of masked entries under modality pair $(m_i{=}1,m_j{=}m)$.

We bucket $n_t$ and compute a $z$-score using bucket-wise statistics $(\mu_b,\sigma_b)$ estimated from a calibration set (falling back to global/neighboring buckets if needed). 
We then determine $m_j^{*}$, the most influential modality for $\text{drug}_j$, by selecting the modality whose masking yields the largest size-calibrated perturbation:

\begin{equation}
m_j^{*}
=\arg\max_{m_j}\;
\frac{d(m_i, m_j)-\mu_{b(n_t)}}{\sigma_{b(n_t)}+\varepsilon}.
\label{eq:cs_stage1}
\end{equation}

Given $m_j^{*}$, we enumerate $t_i\in \mathrm{tails}^{(1)}_{i}$ and $t_j\in \mathrm{tails}^{(m_j^{*})}_{j}$, and perform cell-level masking by occluding only the two entries $(1,\text{drug}_i,t_i)$ and $(m_j^{*},\text{drug}_j,t_j)$ in the relation matrix $\mathbf{R}_{\text{full}}$ (replaced with a reference value, e.g., zero). The tail-pair importance is
\begin{equation}
d(t_i,t_j)=\Big\|f(\mathbf{R}_{\text{full}})-f(\mathbf{R}^{(t_i,t_j)}_{\text{mask}})\Big\|^{2},
\label{eq:cs_stage2}
\end{equation}
and we rank tail pairs by $d(t_i,t_j)$ and report Top-$1$.

\section{Experiments}

\begin{table*}[htbp]
\centering
\caption{The performance of CMF-ELN and its competitors on (known drug)–(new drug) of Dataset 1 and Dataset 2.} 
\label{tab:model_comparison_one} 
\small
\begin{threeparttable}
    \resizebox{\linewidth}{!}{
    \begin{tabular}{cc|ccccccccc}
    \hline\hline
    \textbf{Dataset} & \textbf{Metric} & \textbf{MDDI-SCL} & \textbf{MDF-SA-DDI} & \textbf{MDNN} & \textbf{MKG-FENN} & \textbf{Know-DDI} & \textbf{Emer-GNN} & \textbf{MeT-DDI} & \textbf{ImageDDI} & \textbf{CMF-ELN} \\
    \hline
    \multirow{6}{*}{Dataset 1} 
     & \textbf{ACC} & 0.7011±0.021 & 0.7029±0.018 & 0.6933±0.025 & 0.6625±0.014 & 0.6712±0.027 & 0.6828±0.016 & 0.6872±0.017 & \underline{0.7049}±0.018 & \textbf{0.7276}±0.025 \\
     & \textbf{AUPR} & \underline{0.7632}±0.012 & 0.7437±0.022 & 0.7337±0.017 & 0.7359±0.020 & 0.6248±0.031 & 0.6129±0.029 & 0.7493±0.011 & 0.7523±0.007 & \textbf{0.7977}±0.031 \\
     & \textbf{AUC} & \underline{0.8995}±0.009 & 0.8901±0.013 & 0.8809±0.020 & 0.8848±0.011 & 0.8971±0.019 & 0.8876±0.018 & 0.8875±0.031 & 0.8981±0.011 & \textbf{0.9169}±0.019 \\
     & \textbf{F1} & \underline{0.6939}±0.020 & 0.5993±0.018 & 0.6848±0.016 & 0.6408±0.025 & 0.5209±0.023 & 0.5036±0.019 & 0.6684±0.011 & 0.6825±0.007 & \textbf{0.7267}±0.024 \\
     & \textbf{Pre} & \underline{0.7113}±0.014 & 0.6873±0.027 & 0.7006±0.019 & 0.6728±0.031 & 0.6279±0.017 & 0.6421±0.022 & 0.6921±0.018 & 0.7022±0.012 & \textbf{0.7420}±0.007 \\
     & \textbf{Rec} & \underline{0.6796}±0.023 & 0.5848±0.019 & 0.6718±0.020 & 0.6189±0.021 & 0.5628±0.013 & 0.6721±0.025 & 0.6537±0.022 & 0.6679±0.008 & \textbf{0.7138}±0.015 \\
    \hline
    \multirow{6}{*}{Dataset 2} 
     & \textbf{ACC} & 0.7122±0.017 & 0.7035±0.021 & \underline{0.7146}±0.020 & 0.6632±0.028 & 0.6972±0.015 & 0.6706±0.019 & 0.6659±0.011 & 0.6927±0.017 & \textbf{0.7323}±0.023 \\
     & \textbf{AUPR} & 0.7123±0.022 & 0.7169±0.019 & \underline{0.7326}±0.025 & 0.6451±0.018 & 0.6837±0.023 & 0.6731±0.020 & 0.6945±0.021 & 0.6807±0.009 & \textbf{0.7716}±0.021 \\
     & \textbf{AUC} & 0.9621±0.014 & \underline{0.9760}±0.016 & 0.9624±0.013 & 0.9289±0.020 & 0.9256±0.015 & 0.9353±0.017 & 0.9763±0.012 & 0.9233±0.011 & \textbf{0.9788}±0.014 \\
     & \textbf{F1} & \underline{0.6322}±0.018 & 0.6121±0.015 & 0.6237±0.019 & 0.6106±0.016 & 0.5924±0.020 & 0.6004±0.021 & 0.4727±0.022 & 0.5898±0.021 & \textbf{0.6171}±0.015 \\
     & \textbf{Pre} & 0.7122±0.022 & 0.7036±0.017 & \underline{0.7178}±0.020 & 0.6424±0.019 & 0.6781±0.021 & 0.6521±0.016 & 0.5285±0.008 & 0.6761±0.005 & \textbf{0.7022}±0.026 \\
     & \textbf{Rec} & 0.5519±0.019 & \underline{0.5601}±0.015 & 0.5597±0.021 & 0.5379±0.017 & 0.5231±0.020 & 0.5279±0.019 & 0.4795±0.031 & 0.5325±0.016 & \textbf{0.5738}±0.011 \\
    \hline
    \multirow{3}{*}{Statistic*} 
     & \textbf{Win/Loss} & 10/2 & 11/1 & 10/2 & 12/0 & 12/0 & 12/0 & 12/0 & 12/0 & \textbf{91/5} \\
     & \textbf{\textit{p}-value} & 0.0047 & 0.0029 & 0.0060 & 0.0022 & 0.0022 & 0.0022 & 0.0022 & 0.0022 & \textbf{-} \\
     & \textbf{F-rank} & 2.75 & 4.42 & 3.83 & 7.17 & 7.08 & 6.92 & 6.25 & 5.17 & \textbf{1.42} \\
    \hline\hline
    \end{tabular}
    }
    \begin{tablenotes}[para, flushleft]
        \footnotesize 
        \item[*]\underline{Underlined}: Best baseline value. 
        \textbf{Win/Loss}: Comparison of CMF-ELN's performance across metrics. 
        \textbf{P-value}: Wilcoxon signed-rank test results.\\
        \textbf{F-rank}: average Friedman rank across six metrics. 
    \end{tablenotes}    
\end{threeparttable}
\end{table*}

\begin{table*}[htbp]
\centering
\caption{The performance of CMF-ELN and its competitors on (new drug)–(new drug) of Dataset 1 and Dataset 2.} 
\label{tab:model_comparison_two} 
\small
\begin{threeparttable}
    \resizebox{\linewidth}{!}{
    \begin{tabular}{cc|ccccccccc}
    \hline\hline
    \textbf{Dataset} & \textbf{Metric} & \textbf{MDDI-SCL} & \textbf{MDF-SA-DDI} & \textbf{MDNN} & \textbf{MKG-FENN} & \textbf{Know-DDI} & \textbf{Emer-GNN} & \textbf{MeT-DDI} & \textbf{ImageDDI} & \textbf{CMF-ELN} \\
    \hline
    \multirow{6}{*}{Dataset 1} 
     & \textbf{ACC}  & 0.5533±0.021 & 0.5391±0.018 & 0.5316±0.027 & 0.4758±0.016 & 0.5172±0.020 & 0.5231±0.023 & \underline{0.5581}±0.025 & 0.5506±0.012 & \textbf{0.5827}±0.031 \\
     & \textbf{AUPR} & 0.5404±0.014 & 0.5417±0.017 & 0.5378±0.019 & 0.4671±0.021 & \underline{0.5721}±0.013 & 0.5698±0.015 & 0.5563±0.022 & 0.5497±0.007 & \textbf{0.6194}±0.015 \\
     & \textbf{AUC}  & 0.8022±0.010 & 0.8137±0.015 & 0.7876±0.014 & 0.7511±0.009 & \underline{0.8236}±0.014 & 0.8189±0.014 & 0.7942±0.011 & 0.7896±0.018 & \textbf{0.8321}±0.019 \\
     & \textbf{F1}   & 0.5405±0.018 & 0.5213±0.014 & 0.5087±0.023 & 0.4102±0.020 & 0.5213±0.019 & \underline{0.5509}±0.017 & 0.5157±0.014 & 0.5134±0.007 & \textbf{0.5732}±0.024 \\
     & \textbf{Pre}  & 0.5584±0.022 & \underline{0.6139}±0.018 & 0.5394±0.017 & 0.4876±0.021 & 0.6039±0.015 & 0.6123±0.025 & 0.5521±0.011 & 0.5547±0.006 & \textbf{0.5997}±0.015 \\
     & \textbf{Rec}  & \underline{0.5270}±0.016 & 0.5231±0.020 & 0.4893±0.023 & 0.3888±0.018 & 0.5234±0.021 & 0.5190±0.013 & 0.4982±0.026 & 0.5182±0.013 & \textbf{0.5543}±0.017 \\
    \hline
    \multirow{6}{*}{Dataset 2} 
     & \textbf{ACC}  & 0.5008±0.018 & 0.5105±0.021 & \underline{0.5110}±0.020 & 0.4143±0.022 & 0.4421±0.017 & 0.4674±0.019 & 0.4471±0.016 & 0.4925±0.012 & \textbf{0.5161}±0.021 \\
     & \textbf{AUPR} & 0.4787±0.023 & \underline{0.4926}±0.019 & 0.4598±0.020 & 0.3722±0.021 & 0.4198±0.016 & 0.4134±0.018 & 0.4118±0.013 & 0.4633±0.009 & \textbf{0.4965}±0.023 \\
     & \textbf{AUC}  & \underline{0.9406}±0.015 & 0.9361±0.017 & 0.9049±0.023 & 0.9189±0.014 & 0.9327±0.018 & 0.9376±0.019 & 0.9355±0.016 & 0.9233±0.014 & \textbf{0.9382}±0.018 \\
     & \textbf{F1}   & 0.2752±0.017 & 0.2609±0.018 & \underline{0.2789}±0.019 & 0.1926±0.024 & 0.2203±0.017 & 0.2311±0.022 & 0.2722±0.024 & 0.2774±0.015 & \textbf{0.2830}±0.014 \\
     & \textbf{Pre}  & \underline{0.3128}±0.017 & 0.3124±0.020 & 0.2923±0.021 & 0.2231±0.018 & 0.3023±0.007 & 0.3101±0.016 & 0.3073±0.006 & 0.3031±0.021 & \textbf{0.3558}±0.025 \\
     & \textbf{Rec}  & 0.2483±0.022 & 0.2372±0.017 & 0.2662±0.018 & 0.1868±0.021 & 0.2159±0.019 & 0.2098±0.024 & \underline{0.2871}±0.006 & 0.2661±0.008 & \textbf{0.2569}±0.024 \\
    \hline
    \multirow{3}{*}{Statistic*} 
     & \textbf{Win/Loss} & 11/1 & 11/1 & 11/1 & 12/0 & 11/1 & 11/1 & 11/1 & 11/1 & \textbf{89/7} \\
     & \textbf{\textit{p}-value} & 0.0029 & 0.0060 & 0.0047 & 0.0022 & 0.0028 & 0.0037 & 0.0060 & 0.0037 & \textbf{-} \\
     & \textbf{F-rank} & 3.67 & 4.04 & 6.17 & 8.92 & 5.58 & 4.75 & 5.25 & 5.25 & \textbf{1.58} \\
    \hline\hline
    \end{tabular}
    }
\end{threeparttable}
\end{table*}


In the subsequent experiments, we primarily focus on addressing the following three research questions:

\begin{itemize}
    \item RQ.1: Does the proposed CMF-ELN model outperform the state-of-the-art models in predicting DDIs under two scenarios: between one known drug and one new drug (Task 1), and between two new drugs (Task 2)?
    \item RQ.2: What are the contributions of each modality in the proposed CMF-ELN model (ablation study)?
\end{itemize}
\subsection{General Settings}

\textbf{Datasets.} Two benchmark datasets from DrugBank are employed: Dataset 1 \cite{MeTDDI2024} (1,409 drugs; 343,036 pairs; 4 DDI types) and Dataset 2 \cite{DeepDDI} (1,710 drugs; 192,284 pairs; 86 DDI types). We further integrated four data modalities for each drug, specifically protein, Morgan fingerprints, MACCS keys, and structural motifs.


\textbf{Evaluate Metrics.} For the purpose of assessing the proposed CMF-ELN model, we employ a collection of evaluation metrics suitable for multi-class classification tasks. This set of metrics encompasses accuracy (ACC), the area under the precision-recall curve (AUPR), the area under the receiver operating characteristic curve (AUC), F1 score, precision (Pre), and recall (Rec).

\textbf{Baselines.} We compare CMF-ELN against eight state-of-the-art baselines, including MKG-FENN \cite{MKG-FENN}, KnowDDI \cite{KnowDDI}, Emer-GNN \cite{EmerGNN}, MDF-SA-DDI \cite{MDF-SA-DDI}, MDDI-SCL \cite{MDDI-SCL}, MDNN \cite{MDNN}, MetDDI \cite{MeTDDI2024}, and ImageDDI \cite{He2025ImageDDI}.

\textbf{Implementation details.} In the experimental configuration, the model was trained with an embedding size of 128, a learning rate of $1 \times 10^{-3}$, and a regularization weight of $10^{-6}$.

\subsection{Comparative Performance with Baseline Approaches (RQ.1)}
Tables~\ref{tab:model_comparison_one} and~\ref{tab:model_comparison_two} report the performance comparison between CMF-ELN and the baseline models under the two cold-start settings. Following a five-fold cross-validation protocol, we designate one-fifth of drug types as ``new'' for each fold and isolate all associated DDI pairs to form the test set. Overall, CMF-ELN outperforms the competitors in most cases and yields the best Friedman rank. Specifically, CMF-ELN achieves 91 wins and 5 losses in the known drug--new drug setting (Table~\ref{tab:model_comparison_one}), and 89 wins and 7 losses in the new drug--new drug setting (Table~\ref{tab:model_comparison_two}). Moreover, CMF-ELN achieves the superior Friedman ranks~\cite{friedman1937use} of 1.42 and 1.58 in the two settings, respectively, demonstrating its consistently strong performance over the comparison methods. Furthermore, the calculated $p$-values~\cite{demvsar2006statistical} across all datasets are less than 0.05, confirming that the performance improvements achieved by the proposed CMF-ELN model are statistically significant.

\subsection{Ablation Analysis (RQ.2)}

\begin{table}[!ht]
    \centering
    \scriptsize
    \setlength{\tabcolsep}{3.5pt}  
    \begin{tabular}{@{}cccccccc@{}}  
        \toprule
        \textbf{Modalities} & \textbf{ACC} & \textbf{AUPR} & \textbf{AUC} & \textbf{F1} & \textbf{P} & \textbf{R} & \textbf{F-rank} \\
        \midrule
        1234 & 0.7276 & 0.7977 & 0.9169 & 0.7267 & 0.7420 & 0.7138 & 1.83 \\
        \midrule
        123  & 0.7246 & 0.7951 & 0.9161 & 0.7226 & 0.7418 & 0.7069 & 3.25 \\
        124  & 0.7232 & 0.7883 & 0.9137 & 0.7224 & 0.7414 & 0.7069 & 4.75 \\
        134  & 0.7197 & 0.8036 & 0.9188 & 0.7129 & 0.7468 & 0.6887 & 3.33 \\
        234  & 0.6905 & 0.7351 & 0.8853 & 0.6803 & 0.6977 & 0.6661 & 11.17 \\
        \midrule
        12   & 0.7211 & 0.7959 & 0.9155 & 0.7147 & 0.7538 & 0.6877 & 3.83 \\
        13   & 0.7192 & 0.7931 & 0.9148 & 0.7146 & 0.7517 & 0.6886 & 4.83 \\
        14   & 0.7141 & 0.7973 & 0.9147 & 0.7051 & 0.7356 & 0.6827 & 6.17 \\
        23   & 0.6872 & 0.7481 & 0.8914 & 0.6733 & 0.7141 & 0.6465 & 9.50 \\
        24   & 0.6851 & 0.7441 & 0.8882 & 0.6691 & 0.7084 & 0.6431 & 11.33 \\
        34   & 0.6792 & 0.7444 & 0.8895 & 0.6623 & 0.6965 & 0.6389 & 12.17 \\
        \midrule
        1    & 0.7011 & 0.7808 & 0.9084 & 0.6968 & 0.7295 & 0.6735 & 8.00 \\
        2    & 0.6843 & 0.7403 & 0.8867 & 0.6709 & 0.7068 & 0.6464 & 11.67 \\
        3    & 0.6759 & 0.7394 & 0.8885 & 0.6612 & 0.6886 & 0.6412 & 13.17 \\
        4    & 0.6516 & 0.7101 & 0.8734 & 0.6317 & 0.6624 & 0.6104 & 15.00 \\
        \bottomrule
    \end{tabular}
    \vspace{2mm}
    \caption{Ablation study of CMF-ELN cross-modal knowledge graph components on Dataset 1 under Task 1.}
    \label{tab:mkg_comb1}
\end{table}

To systematically investigate the impact of various drug embedding combinations from distinct channels, comprehensive ablation studies were conducted to assess modality contributions. Specifically, the numbers used to denote CMF-ELN variants correspond to the four modality-specific knowledge graph channels: 1 represents the Drug-Protein channel, 2 represents the Drug-Morgan channel, 3 represents the Drug-MACCS channel, and 4 represents the Drug-Motif channel. Therefore, variants 1--4 denote single-channel settings, whereas combinations such as 12, 123, and 1234 denote multi-channel fusion settings. Multiple CMF-ELN variants were developed, with their predictive performance rigorously assessed via evaluation metrics. The representative results for Dataset 1 are summarized in Table \ref{tab:mkg_comb1}.
Our analysis indicates that synthesizing drug representations from integrated molecular and biological profiles is a highly effective strategy. While single-channel models yielded satisfactory performance, implementing multi-channel fusion architectures led to progressive and significant improvements, with the complete CMF-ELN model achieving optimal results. This enhanced performance is attributed to the strategic integration of diverse channels, which facilitates the exploration of complementary drug facets—ranging from molecular topology and functional groups to biological interactions. By capturing these multifaceted characteristics, the CMF-ELN model gains a more nuanced understanding of complex drug mechanisms, ultimately resulting in the highest accuracy for DDI prediction.

\begin{figure*}[!ht]
    \centering
    \begin{overpic}[width=1.0\textwidth]{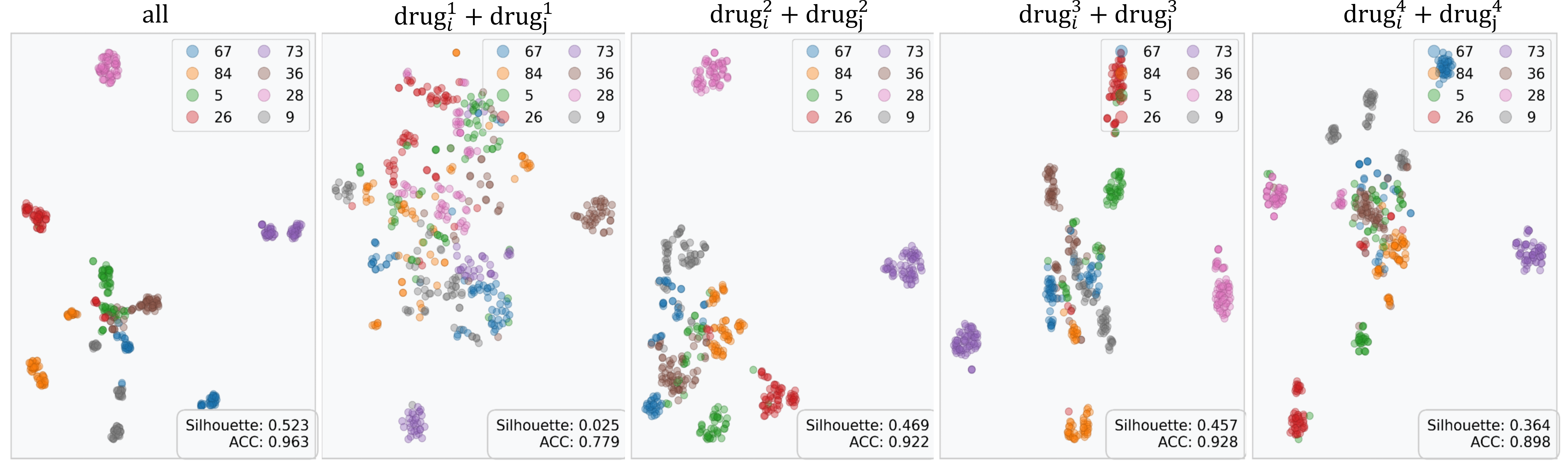}
    \end{overpic}
    \caption{Representation analysis of CMF-ELN. t-SNE visualization of drug-pair embeddings for Task~1 on Dataset 2, comparing four unimodal representations with the full-modality fused representation.}
    \label{fig:tsne_visual} 
\end{figure*}

Figure~\ref{fig:tsne_visual} visualizes the fused and unimodal embeddings for Task~1 on Dataset~2. The t-SNE plots show that the fused embedding yields a more compact intra-class structure and clearer inter-class separation than any unimodal embedding. To quantify this gain, we report the Silhouette coefficient~\cite{peter1987graphical} and KNN accuracy~\cite{cover1967nearest} on the 2D embeddings, providing an objective assessment of clustering quality and label separability.


\section{Conclusion}
In this work, we introduce CMF-ELN, a cross-modal end-to-end framework for cold-start DDI prediction by fusing drug-centered knowledge graphs with multi-view molecular and biomedical evidence. Extensive experiments on two real-world datasets demonstrate that CMF-ELN consistently outperforms state-of-the-art baselines and achieves strong performance in cold-start settings by unifying similarity learning and DDI prediction within a single pipeline. Importantly, the proposed two-stage interpretability scheme (modality identification followed by feature attribution) provides mechanistically coherent evidence chains for both perpetrator and victim drugs, which enhances the practical utility of model outputs by supporting mechanism verification, risk assessment, and targeted experimental or clinical follow-up.

\begin{acks}
This work was supported by the National Natural Science Foundation of China (62576289 and 62302030), the New Chongqing Youth Innovation Talent Project under grant CSTB2024NSCQQC-XMX0035, the Open Project of Key Laboratory of Knowledge Engineering with Big Data at the Ministry of Education of China (BigKEOpen2025-03), and the Southwest University Graduate Student Research Innovation (SWUS26112). Thanks to the High Performance Computing (HPC) clusters at Southwest University. Yi He was not supported by any of the above funding.
\end{acks}

\bibliographystyle{ACM-Reference-Format}
\bibliography{references-arxiv}

\end{document}